\documentclass[letterpaper, 10 pt, conference]{ieeeconf}  % Comment this line out if you need a4paper

\IEEEoverridecommandlockouts                              % This command is only needed if 
\usepackage{graphics} % for pdf, bitmapped graphics files
\usepackage{graphicx}
\usepackage{booktabs}
\usepackage{multirow}
\usepackage{caption}
\usepackage{titlesec} 
\usepackage{hyperref}
\usepackage{subcaption}
\usepackage{cite}
\usepackage{soul}
\usepackage[normalem]{ulem}
\hypersetup{
    colorlinks=true,
    linkcolor=blue,
    filecolor=magenta,      
    urlcolor=cyan,
    pdftitle={Overleaf Example},
    pdfpagemode=FullScreen,
    }

\usepackage{caption}
\usepackage{xcolor}

\title{\LARGE \bf
Pre-Trained Masked Image Model for Mobile Robot Navigation}

\author{Vishnu Dutt Sharma, Anukriti Singh, Pratap Tokekar
\thanks{This work is supported by NSF under grant No. 1943368, ONR under grant number N00014-18-1-2829, and an Amazon Research Award.}
\thanks{The authors are with the Dept. of Computer Science,
        University of Maryland, College Park, MD, USA
        {\tt\small \{vishnuds, anukriti, tokekar\}@umd.edu}}%
}

\begin{document}

\maketitle
\thispagestyle{empty}
\pagestyle{empty}

%%%%%%%%%%%%%%%%%%%%%%%%%%%%%%%%%%%%%%%%%%%%%%%%%%%%%%%%%%%%%%%%%%%%%%%%%%%%%%%%
\begin{abstract}
2D top-down maps are commonly used for the navigation and exploration of mobile robots through unknown areas. Typically, the robot builds the navigation maps incrementally from local observations using onboard sensors. Recent works have shown that predicting the structural patterns in the environment through learning-based approaches can greatly enhance task efficiency. While many such works build task-specific networks using limited datasets, we show that the existing foundational vision networks can accomplish the same without any fine-tuning. Specifically, we use Masked Autoencoders, pre-trained on street images, to present novel applications for field-of-view expansion, single-agent topological exploration, and multi-agent exploration for indoor mapping, across different input modalities. Our work motivates the use of foundational vision models for generalized structure prediction-driven applications, especially in the dearth of training data. We share more qualitative results at \url{https://raaslab.org/projects/MIM4Robots}.

\end{abstract}

\section{INTRODUCTION}

Mobile robot navigation through unknown areas has been studied by the robotics community for a long time~\cite{kuipers2017shakey}. Generally, in the existing approaches, the robot updates the map based on its observations so far and moves according to the task at hand, such as PointGoal navigation, ObjectGoal navigation, and exploration~\cite{anderson2018evaluation}. In the case of ground robots, the map is typically represented in a top-down view or Bird's Eye View (BEV), as the robot motion is constrained on the ground plane. Aerial robots also use Top-down representations for navigation and exploration or to help others when working in tandem with other aerial or ground robots~\cite{albani2017field, sharma2023d2coplan, hood2017bird, sharma2020risk}.  

The traditional approach is to build a map by fusing the robot's observations. Recent works across the wider robotics community have started exploring learning-based approaches to augment the robot's onboard data about the environment, e.g., occupancy map, point cloud, etc., to accomplish tasks~\cite{katyal2018occupancy, katyal2019uncertainty, saroya_online_2020, dhami2023prednbv, sharma2023proxmap}. These methods learn the patterns in representations and can predict the yet-unobserved regions based on partially observed environments. The prediction can then be used to make informed decisions for safer and more efficient motion planning.

\begin{figure}[ht!]
\vspace{3mm}
  \centering
%   \fbox{\rule{0pt}{2in} \rule{0.9\linewidth}{0pt}}
  \includegraphics[width=0.99\linewidth]{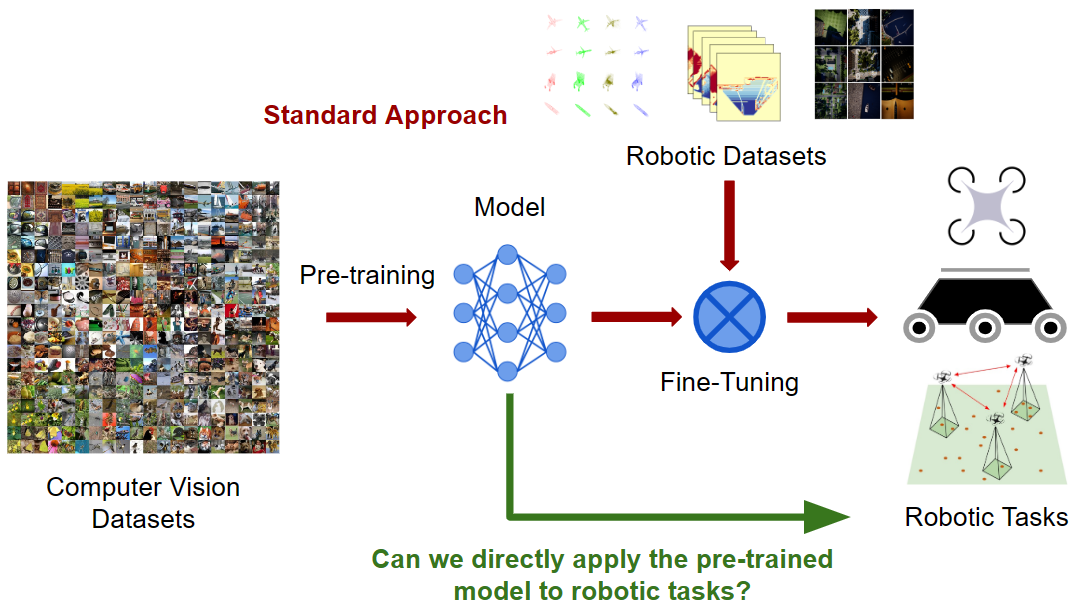}

  \caption{Traditional approach of leveraging models trained on huge computer vision datasets can be applied to robotic tasks reliant on top-down images, albeit with some task-specific fine-tuning. We show that this is not necessary and some models, such as MAE~\cite{he2022masked} can be applied directly to these robotics tasks.}
  \label{fig:overview}
\end{figure}

Learning-based methods require extensive datasets, which are challenging to get in robotics applications compared to computer vision. Simulators can generate virtual data, but face a sim2real gap. Many computer vision methods trained on large datasets may not directly apply to robot applications due to distributional differences in image representation; computer vision datasets are mainly comprised of first-person views, captured from a height often taller than the camera mounted on ground robots. Fine-tuning often seems to be the solution but requires similarity between pre-training and fine-tuning tasks, which can be challenging for robot navigation representations such as top-down images, semantic maps, and occupancy maps.

The recent emergence of self-supervised foundational models, which are trained on huge datasets, aims to achieve generalizability by leveraging a diverse distribution of datasets. This approach is premised on the belief that it should prompt the model to reason about fundamental concepts such as shapes and textures. However, the datasets used may not necessarily include the same distribution of images that we expect to observe during robot navigation.

This raises the question: \emph{Can we apply pre-trained computer vision models directly on robotics tasks such as navigation and exploration without fine-tuning?} Surprisingly, the answer is \ul{yes}. We substantiate this assertion with a masked image model that learns to reconstruct an image using representation learning. Specifically, Masked Autoencoder (MAE)~\cite{he2022masked}, which randomly patches the image to learn local correlation and reconstruct the masked parts. We show how, despite being trained on first-person view ~\cite{deng2009imagenet} images, it can make reasonable predictions about the unseen areas in \textit{top-down} RGB, semantic, and occupancy maps, which improves 2D planning for efficient robot navigation. We find that there is no need to fine-tune MAE on specific tasks for improvement, making it further appealing for robotics applications that may not have adequate training data.

Specifically, we make the following contributions in this paper:
\begin{itemize}
    \item We study MAE as an expainting network for top-down images across RGB, semantic maps, and binary maps, and present quantitative and qualitative results across various degrees of increasing field-of-view for indoor and outdoor images.
    \item We present a novel uncertainty-driven exploration method for 2D semantic map reconstruction using MAE and compare it to non-predictive approaches to highlight the benefits of structural pattern prediction.
    \item We show that MAE can be effectively applied for a case study of single robot navigation aided by occupancy prediction, resulting in more efficient operation compared to a standard, non-predictive baseline method.
    
\end{itemize}

\begin{figure}[t]
\vspace{3mm}
  \centering
%   \fbox{\rule{0pt}{2in} \rule{0.9\linewidth}{0pt}}
  \includegraphics[width=0.99\linewidth]{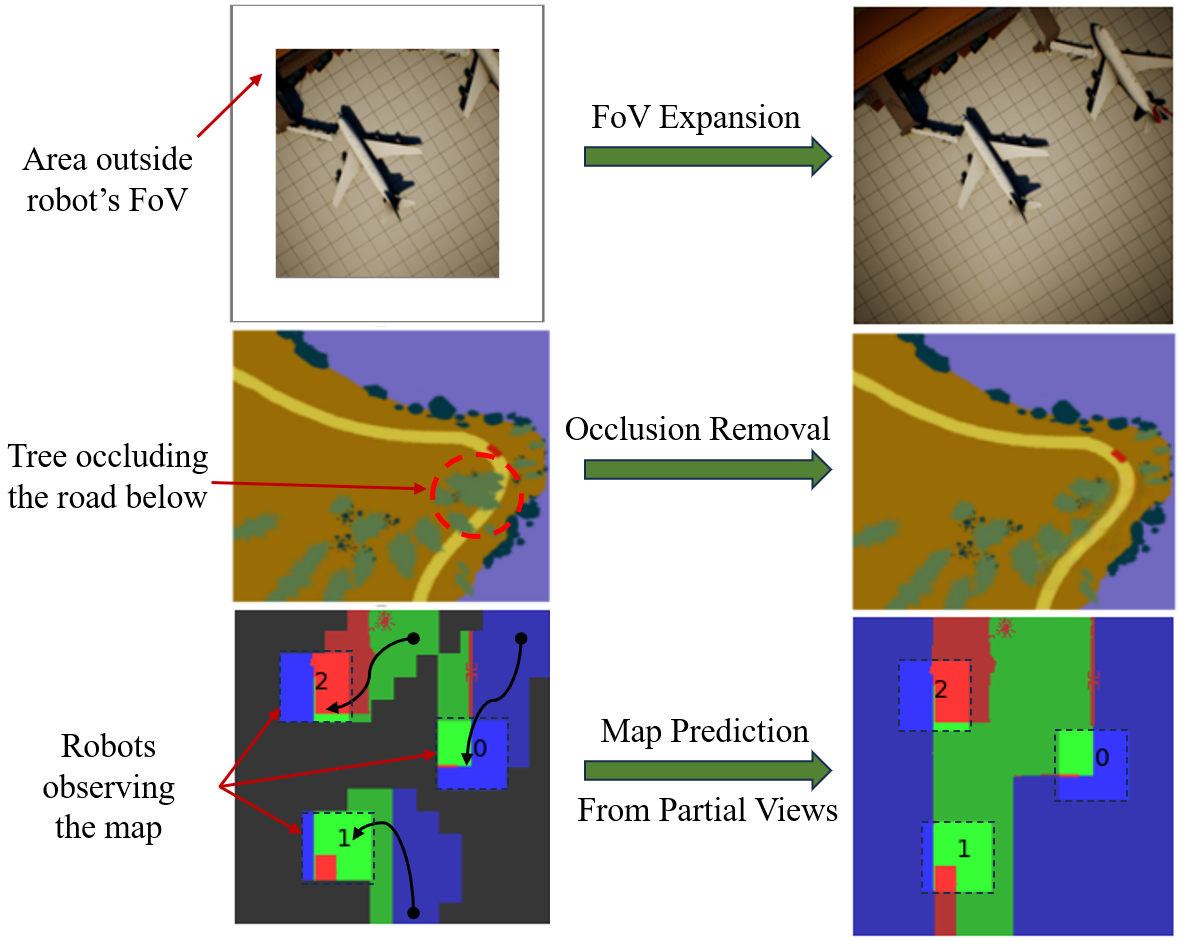}

  \caption{Example of robotics tasks solved with help of Masked Autoencoder.}
  \label{fig:task_examples}
\end{figure}

Our work highlights how foundational self-supervised learning algorithms like masked image model (MAE) can be used for robot tasks by choosing appropriate modalities without any fine-tuning, and paves the way for further improvement to the existing capabilities by task-specific tuning of these models. Coupled with its applicability to a variety of robotics applications, as shown in Fig.~\ref{fig:task_examples}, MAE could potentially be the free-lunch all-around solution for 2D map-based navigation.

\section{Related Works}
\subsection{Mapping for Robot Navigation}
% \subsection{Top-Down Images and FoV expansion for Robot Navigation}
Top-down images and map representations are vital for robot navigation and exploration. Navigating through an unknown map by Simultaneous Mapping and Localization (SLAM), which utilizes the robot's past observations, has been a cornerstone of robotics for robotics. A top-down semantic map is another representation of interest for robotic applications. These maps are useful for semantic goal navigation~\cite{georgakis2021learning, georgakis2022cross}. Top-down images are also beneficial for aerial robot tasks such as surveying and scouting~\cite{del2021unmanned, mohamed2020unmanned}. The maps obtained by the aerial robots can be used to help the ground robots navigate. Semantic maps are obtained from such images to identify navigable and non-navigable areas for the ground robot.

Recent works in this domain have sought to improve task efficiency by \textit{predicting} the unobserved regions of the map to plan ahead. 2D Occupancy map, a top-down representation, has been the focus of many of these works, showing improvement in navigation, exploration distance, and time ~\cite{katyal2018occupancy, ramakrishnan2020occupancy, wei2021occupancy, sharma2022occupancy}. Katyal et al.~\cite {9561034} show these benefits for high-speed navigation, highlighting the importance of predictions. While the predictions are limited to the perception module, it can also enhance planning by extracting uncertainty from the predictions~\cite{katyal2019uncertainty, georgakis2022uncertainty}. The idea of uncertainty extraction also proves helpful in heterogeneous robot teams for risk-aware planning~\cite{sharma2020risk}. The key challenge with all these systems is that they need to be trained on the appropriate modalities, for which sufficient data may not be available, leading us to ask if there exist pre-trained models that can be used in these applications without much training effort, or better, without any fine-tuning at all?

\subsection{Self-supervised masked encoding}

In recent times, various approaches like BEiT~\cite{bao2021beit}, iBOT~\cite{zhou2021ibot}, and ADIOS~\cite{shi2022adversarial} have drawn inspiration from masked language models. These methods have demonstrated remarkable competitiveness in the realm of self-supervised learning (SSL). All three techniques leverage vision transformers and propose strategies to "inpaint" images that have been partially obscured by random masks in various ways. The idea of map prediction is similar to this, and existing works for robotic applications rely on generative models~\cite{koh2022simple, ren2022look, rombach2021geometry}, which require training or fine-tuning networks on simulation data to get accurate results.

Masked Autoencoder (MAE) uses Vision Transformer (ViT) encoder~\cite{dosovitskiy2020image} and is trained to use only the visible patches of an image to predict the missing patches, similar to the training strategy of BERT~\cite{devlin-etal-2019-bert}. MAE uses linear projections and position encodings for feature representation and is trained with mean squared error (MSE) between the reconstructed and original images in the pixel space, but only for masked patches. While MAE is also trained on RGB images only from the ImageNet-1K dataset~\cite{deng2009imagenet}, the underlying ViT architecture allows it to reason about other modalities as shown by MultiMAE~\cite{bachmann2022multimae}. Therefore, we use MAE for our study and show its effectiveness for prediction and inpainting across various modalities in top-down images useful for robotic tasks, without any finetuning.

 \begin{figure}[t]
 \vspace{3mm}
  \centering
%   \fbox{\rule{0pt}{2in} \rule{0.9\linewidth}{0pt}}
  \includegraphics[width=0.99\linewidth]{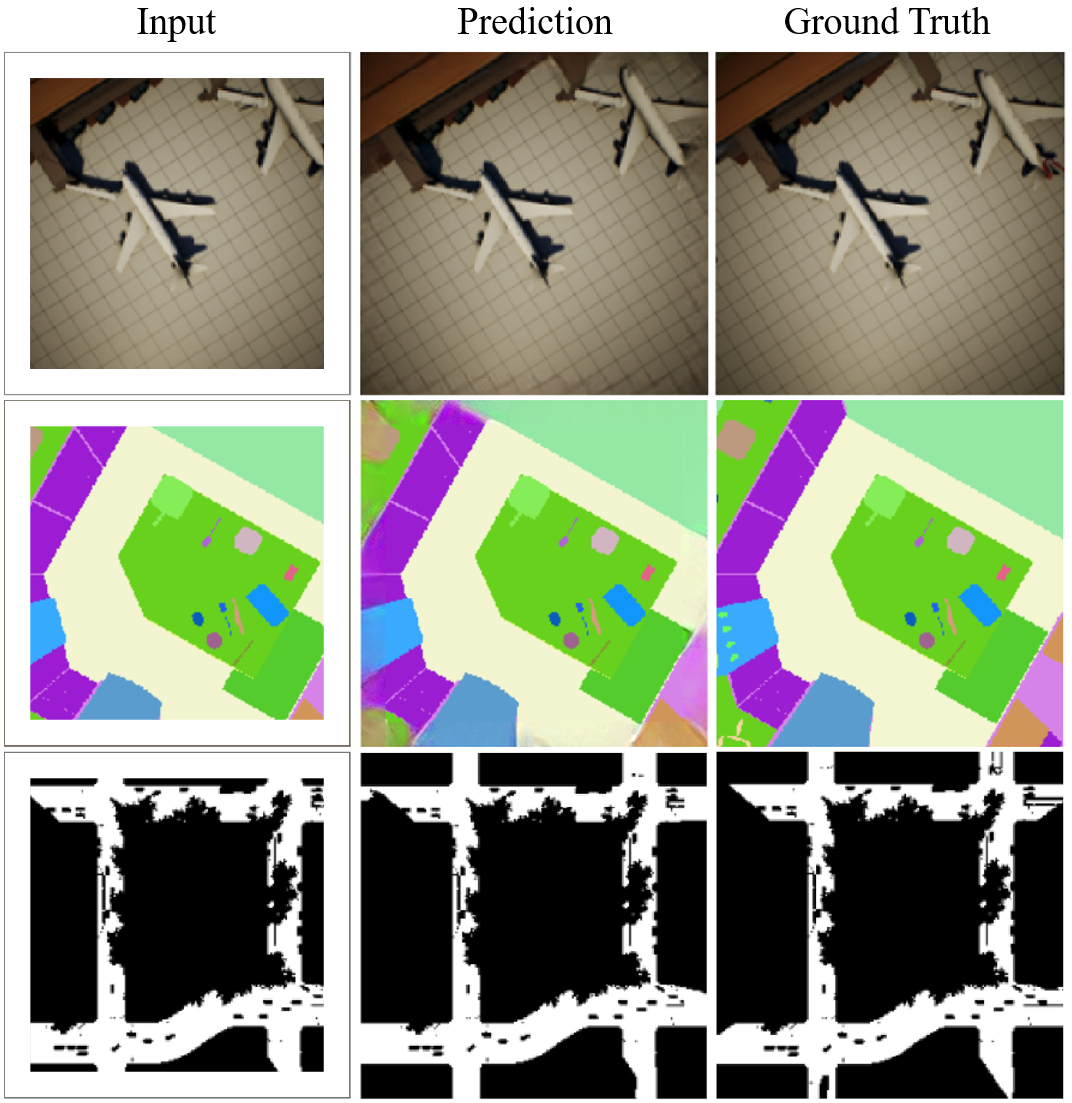}

  \caption{Masked Autoencoder can be used to expand the effective FoV in top-down RGB, semantic, and binary images without fine-tuning.}
  \label{fig:fov_all_modal}
  % \vspace{-2mm}
\end{figure}

\begin{figure}[t]
\vspace{3mm}
  \centering
%   \fbox{\rule{0pt}{2in} \rule{0.9\linewidth}{0pt}}
  \includegraphics[width=0.99\linewidth]{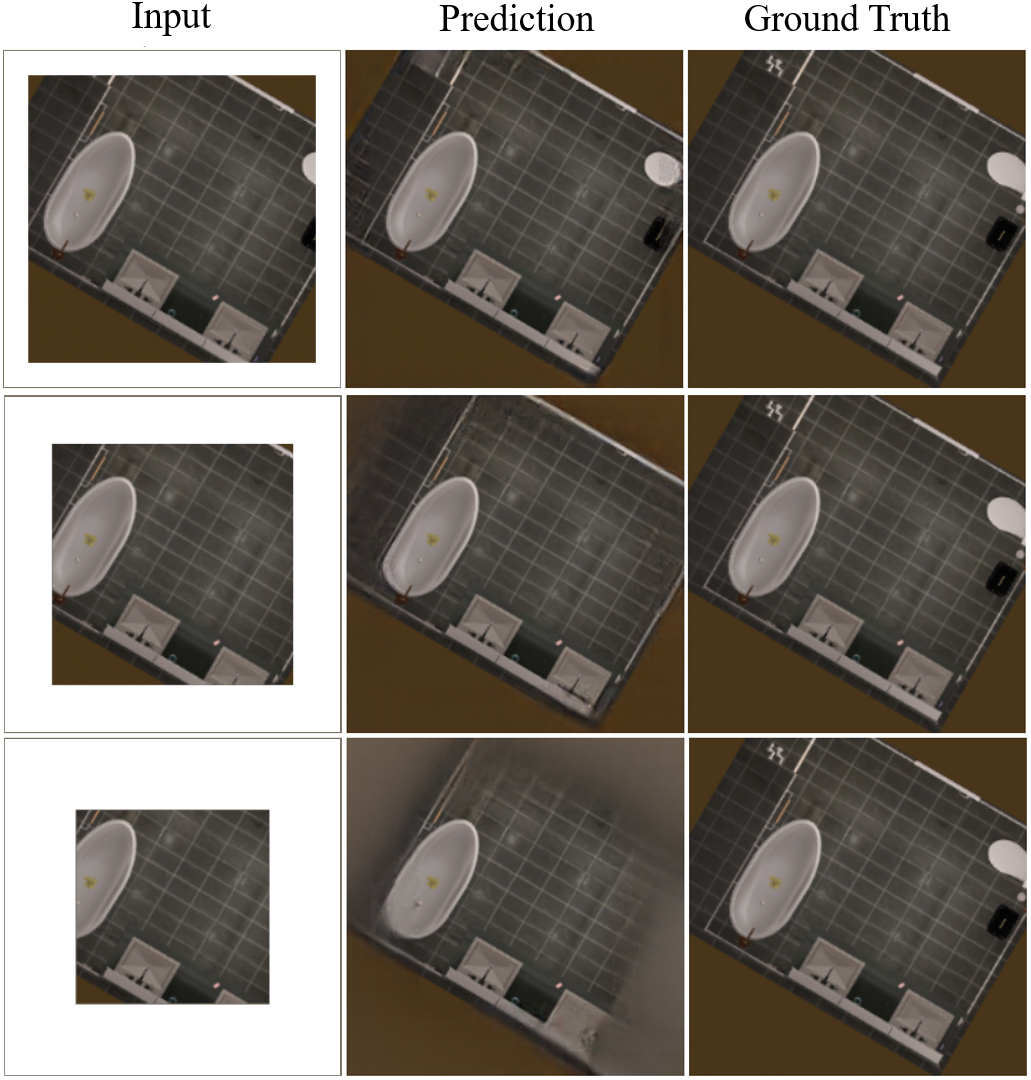}

  \caption{Results of expanding FoV for indoor images in three masking scenarios. The corner of the bathtub and room is accurately predicted based on the symmetry of the lines.  }
  \label{fig:fov_indoors}
  % \vspace{-3mm}
\end{figure}

\section{Methodology}
\label{sec:method}
We aim to determine whether the pre-trained masked autoencoder can effectively predict unobserved regions on 2D maps, represented as top-down RGB images, semantic maps, or binary maps. We focus on three tasks relevant to robot navigation and exploration, with detailed descriptions provided in the following subsections.

\subsection{FoV Expansion and Navigation}
\label{subsec:method_fov_expansion}
Katyal et al.~\cite{katyal2018occupancy} conducted a comprehensive study on various convolutional networks to augment the effective Field of View (FoV) of the robot for predicting unexplored occupancy maps in the robot's surroundings. In their subsequent work, they demonstrated that the prediction of future occupancy maps can improve high-speed navigation ~\cite{katyal2019uncertainty}. This research employed U-Net~\cite{ronneberger2015u} as an image-to-image translation network for occupancy map prediction, founding the basis of subsequent research to further enhance robot navigation and exploration~\cite{ramakrishnan2020occupancy, wei2021occupancy, georgakis2022uncertainty, sharma2023proxmap}.

In this study, we primarily investigate the FoV expansion task, as shown in Fig.~\ref{fig:fov_all_modal} and~\ref{fig:fov_indoors}. Instead of employing raw occupancy maps, we opt for semantic segmentation maps and binary maps, modalities that are eventually used by conventional robotic planners. Additionally, we study RGB images, a modality consistent with the one used for MAE training and relevant to aerial mapping and surveying applications. This allows us to examine (a) whether MAE can work well on a different camera view, and (b) how other modalities, i.e., semantic and binary maps, perform during inference when compared with the one used for training MAE. Furthermore, we extend the original study by evaluating MAE performance in both indoor and outdoor environments. The inputs to MAE are provided as 3-channel images, with labels replaced by corresponding colors in semantic and binary maps. Subsequently, the colors in the output images are reconverted to labels by substituting them with the label associated with the closest color in the input images. These labels are then utilized as the assigned classes for evaluation.

\subsection{Multi-Agent Uncertainty Guided Exploration}
\label{subsec:method_multi_agent_exploration}
Uncertainty-guided navigation and exploration, as proposed in previous works~\cite{katyal2019uncertainty, sharma2020risk, georgakis2022uncertainty} aims to enhance active robot exploration by combining the uncertainty-driven exploration technique with image inpainting networks. The eventual goal is to efficiently map the whole environment. These tasks, however, limit themselves to single-agent applications. We propose a novel approach along these lines for a multi-agent setup with pre-trained MAE, without any architectural changes such as dropout injection~\cite{gal2015dropout}.

Our approach draws on concepts from bootstrapping~\cite{efron1992bootstrap} and adversarial attacks on neural networks~\cite{szegedy2014intriguing}. By injecting minimal random noise into the input image, we obtain predictions on perturbed inputs from MAE. Despite resulting in imperceptible visual changes. We repeat this procedure multiple times to get $n$ predictions on such \textit{bootstrapped} inputs from MAE and find variance across each pixel, summed over the channels, as the uncertainty in prediction. Conceptually, pixels with high variance indicate regions where MAE lacks strong structural cues from visible input, necessitating direct observations from the robot. 

\begin{figure}
  \centering
  \vspace{3mm}
  \includegraphics[width=0.99\linewidth]{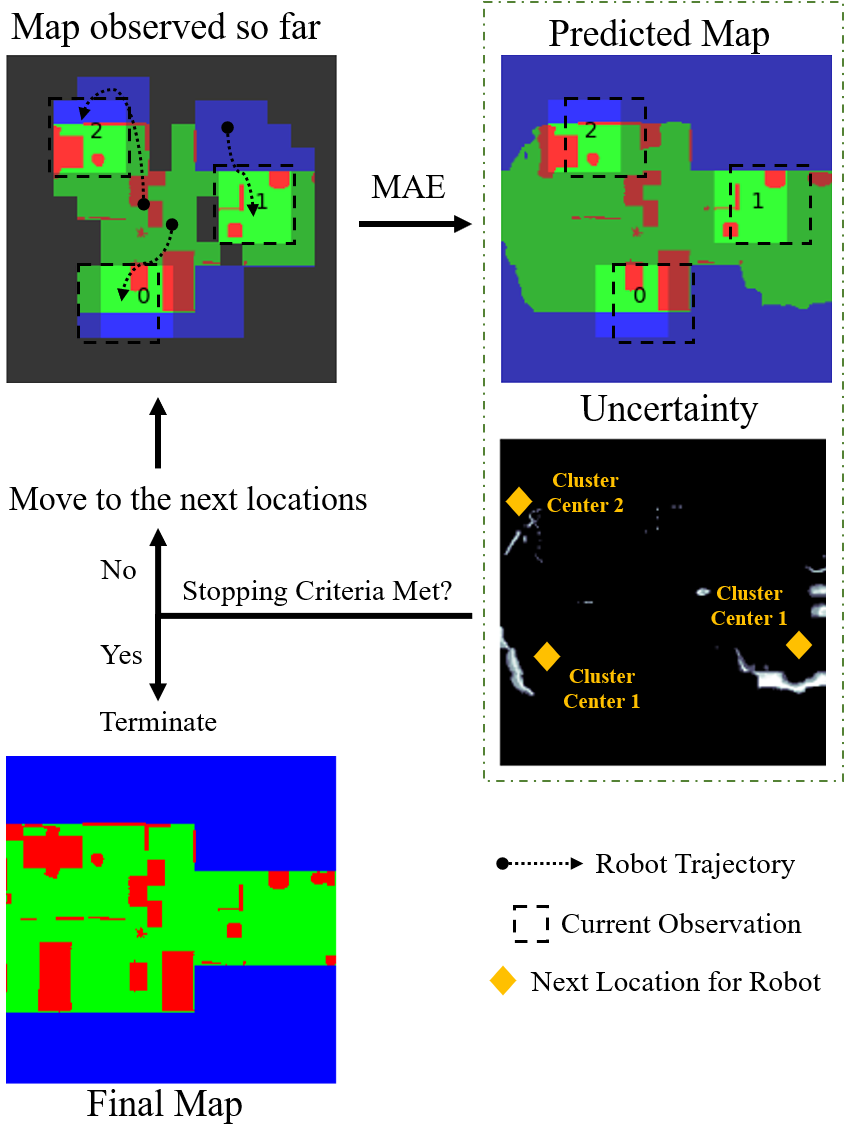}
  \caption{An overview of the multi-agent exploration pipeline.}
  \label{fig:multiagent_explore_overview}
  \vspace{-3mm}
\end{figure}

We put this premise to the test by looking at the prediction accuracy at each step of the exploration. To execute exploration, unexplored locations and those with high uncertainty are subsequently grouped together to identify distinct regions for potential exploration. The robots are assigned to this cluster based on their proximity to the cluster center while ensuring that no two robots are assigned to the same region. We stop the exploration when the cluster centers stabilize. Fig.~\ref{fig:multiagent_explore_overview} shows an overview of this process. This study addresses two critical questions: (a) how to extract uncertainty from MAE, a point-prediction network, and (b) can predictions be leveraged to fill gaps in unexplored maps when resource constraints, such as battery limitations for aerial robots, hinder complete coverage?

We compared the following algorithms for this task: 
\begin{itemize}
    \item  \textbf{Boustrophedon Cell Decomposition Algorithm (Lawnmower)}: Proposed by Choset et al.~\cite{choset1998coverage}, this algorithm divides the regions into $n$ contiguous scanlines of similar size, each assigned to one robot. Each robot scans the designated area for coverage. For this method, we position the robots at the start of the respective scanlines to streamline the process.

    \item \textbf{KMeans Clustering (KMeans-U)}: Here, we apply the KMeans algorithm (with $n$ centers) to the Cartesian coordinates of the unexplored grid cells to identify the centers of the unexplored regions. The robots are then assigned to these regions based on proximity to the cluster and move towards them. At each step, the robots observe the region below and include it in the known map. Then we repeat the clustering process to find centers for the remaining unexplored areas.

    \item \textbf{KMeans Clustering followed by Reconnaissance (KMeans-R)}: Employing KMeans directly may lead to unexplored regions at the center of the map when the cluster centers stabilize (which is a stopping criterion). To address this, we introduce an additional step of relocating all the robots to the center of the map after stabilization. This results in enhanced coverage at the expense of time.

    \item \textbf{KMeans Clustering on Unknown and Uncertain Regions (KMeans-U$^2$)}: In this method, we extract uncertainty from MAE and use the locations with non-zero variance, along with those yet unexplored, for clustering. The procedure for assigning clusters to robots follows a similar approach as in the earlier methods.
\end{itemize}

Here we aim to assess the prediction capabilities of MAE and thus make predictions on the map explored so far at each step. We compare the distributions of coverage to reach $95\%$ prediction accuracy to find which algorithm is more efficient in predicting the unexplored map. 

%%%%%%%%%%%%%%%%%%%%%%%%%%%%%%%%%%%%%%%%%%%%%%%%%
%%%%%%%%%%%%%%%%%%%%%%%%%%%%%%%%%%%%%%%%%%%%%%%%%%

\subsection{Navigation with Prediction}
\label{subsec:method_navigation}
For autonomous navigation, it is crucial to know the map of the environment. The classical methods treat the unexplored area as unknown and build a costmap on the basis of only the current observation. 
The robot can traverse to the edge of the frontier before needing another observation to plan the path ahead. Effectively, the sensor range of the robot defines the maximum distance it can traverse at once.
 
Previous works have shown that predicting future occupancy can result in faster navigation~\cite{katyal2021high} and smoother control~\cite{elhafsi2020map}. These works train neural networks to make these predictions, using synthetically generated data and real-world data obtained by running a robot around. While the former may run into Sim2Real issues, the data collection with the latter is an arduous process. We test if pre-trained MAE could instead be used for prediction while side-stepping the data issues.

\begin{figure}[t]
\vspace{3mm}
  \centering
%   \fbox{\rule{0pt}{2in} \rule{0.9\linewidth}{0pt}}
  \includegraphics[width=0.98\linewidth]{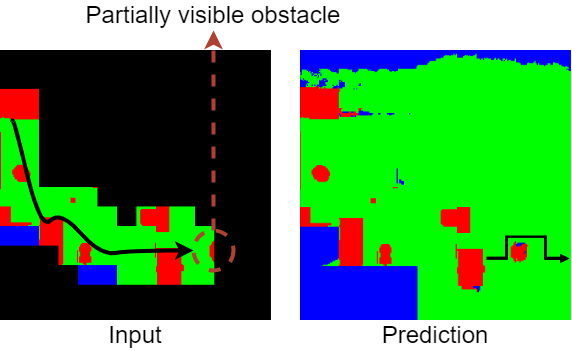}

  \caption{Left: The area robot has explored till now. Right: Prediction of obstacle (red) shape aiding robot path planning.}

  \label{fig:fig_nav}
  \vspace{-3mm}
\end{figure}

For this, we use the predictions of MAE with a standard path planning algorithm on multiple 
indoor floor plans. The robot starts from its initial position where the rest of the area in the map is hidden; using MAE we reconstruct the unseen map and update the path at every next step. The predicted unseen map acts as an estimate of the occupancy ahead, which helps to reconstruct an informed costmap for navigation.
% the increased FoV for the robot which helps to reconstruct a better costmap for navigation. 
This helps the robot cover a greater distance at once by moving to the frontier of every step. 
% Compared to the traditional method, where the robot only knows the current FoV and takes a small step at a time, our method can take a larger step and reach the goal more efficiently. 
Figure~\ref{fig:fig_nav} shows one such example where the robot is able to plan a shorter path ahead since the predictions help to know the shape of the obstacle before the robot actually explores that area. 

We compare our prediction-based approach with a non-predictive approach and calculate the number of steps (and observations) needed to reach a pre-defined goal.
%%%%%%%%%%%%%%%%%%%%%%%%%%%%%%%%%%%%%%%%%%%%%%%%%
%%%%%%%%%%%%%%%%%%%%%%%%%%%%%%%%%%%%%%%%%%%%%%%%%%

\section{Experimental Setups and Evaluation}
In this section, we describe the experimental setup and our findings for each task defined in Section~\ref{sec:method}. Throughout our experiments, we utilize the MAE based on ViT-Large trained on ImageNet-1K dataset~\cite{deng2009imagenet}. 

\subsection{FoV Expansion}
\label{subsec:eva_fov_expansion}

To study FoV expansion, we mask the periphery of the given image by different amounts. MAE uses patches of size $16\times16$ pixels, and masking a patch requires all the pixels in the patch to be masked. We mask the images with 1-3 patches on each side, resulting in an expansion of \textbf{1.17x}, \textbf{1.4x}, and \textbf{1.75x} to the robot's perceptual range, i.e., the number of pixels in a direction if the robot is at the center. 
As the data used by Katyal et al.~\cite{katyal2018occupancy} is not publicly available, we perform an evaluation on the dataset collected from two photorealistic simulated environments, consisting of diverse indoor and outdoor scenes. 

\textbf{Indoor Data:} For indoor environment, we use AI2-THOR~\cite{kolve2017ai2} which has 120 indoor scenes such as kitchens, living rooms, bathrooms, etc. We collect 1444 RGB and segmentation images with a top-down camera of a field of view of 80 degrees and rotated at intervals of 30 degrees (e.g., 30, 60, 90, etc.). 

\textbf{Outdoor Data:} For outdoor images were taken from AirSim VALID dataset ~\cite{9197186} which consist of scenes from cities, suburbs, and mountains among others, captured at different altitudes from an aerial robot. We sample 1000 images from this dataset for this study. For these environments, we also evaluate MAE on binary images, consisting of navigable and non-navigable regions, as a stand-in for occupancy maps.

We evaluate the RGB predictions for the FoV increase on the following metrics typically used to quantify visual similarity: (1) Frechet Inception Distance (FID), (2) Structural Similarity Index Measure (SSIM), (3) Peak Signal-to-Noise Ratio (PSNR), and (4) Mean Squared Error (MSE). For the semantic and binary images, we use mean Intersection-over-Union (mIoU) as the key metric but also provide the results for some of the aforementioned metrics since we use MAE to predict visually similar images for these modalities.

\textbf{Results:} Table~\ref{tab:rgb} summarizes the results for RGB images for both types of environments. We find that increasing the FoV results in worse results than expected since MAE, an inpainting network can not reliably predict the outside areas without much context. 1.75x expansion is the extreme case where the predictions get blurry. Figure~\ref{fig:fov_all_modal} and Figure~\ref{fig:fov_indoors} show some examples in RGB outdoor and indoor scenes respectively and highlight this effect. 

Table~\ref{tab:semantic} and Table~\ref{tab:binary} summarize results for semantic and binary maps. The mIoU is very high for 1.17x expansion and goes down with increasing FoV. The effect is worse indoors as it contains many more classes (270) compared to outdoors (30) and thus may not reliably perform color-to-label matching. Also, small objects are within the scene and on the periphery, and MAE can not expand them without seeing some part of them. Note that the mIoU here is not weighted by the labels' population size. Predictions on binary maps are relatively more robust since the size of objects in each class and the difference in color mapping are larger than the semantic maps. These results present an encouraging picture for a network that was not trained on such images. We note that Katyal et al.~\cite{katyal2018occupancy} report a maximum SSIM of 0.523, 0.534, and 0.504 on real-world data for similar expansion factors. MAE results in better SSIM on both semantic segmentation and binary maps in comparison. Katyal et al.~\cite{katyal2018occupancy} report higher numbers, 0.899, 0.0818, and 0.760, on synthetic data which is similar to the distribution used for training their network. We find MAE on semantic segmentation maps still achieves higher SSIM. However, MAE with binary maps do not achieve similar performance, but still produce good results despite being trained on a different modality and camera view.

\begin{table}[h!]
  \centering
    \caption{Results for increasing the FoV in RGB images}
    \begin{tabular}{lc|cccc}
        \toprule
         \textbf{Setup} & \textbf{Expansion} & \textbf{FID $\downarrow$} & \textbf{SSIM $\uparrow$} & \textbf{PSNR $\uparrow$} & \textbf{MSE $\downarrow$}  \\ 
        \midrule
        \multirow{2}*{Indoor} & 1.17x & \textbf{17.83} & \textbf{0.94} & \textbf{27.76} & \textbf{13.76} \\
        {} & 1.40x & 41.79 & 0.86 & 22.23 & 32.42 \\
        % \addlinespace[0.1cm]
        {} & 1.75x & 76.59 & 0.78 & 19.18 & 52.98 \\
        \midrule
        \multirow{2}*{Outdoor} &  1.17x & \textbf{53.66} & \textbf{0.84} & \textbf{26.38} & \textbf{33.59} \\
        {} &  1.40x & 77.91 & 0.69 & 22.79 & 49.91 \\
        % \addlinespace[0.1cm]
        {} &  1.75x & 116.09 & 0.55 & 19.98 & 67.80 \\
        \bottomrule
    \end{tabular}
    \vspace{-3mm}
  \label{tab:rgb}
\end{table}

\begin{table}[h]
  \centering
    \caption{Results for increasing the FoV in Semantic segmentation images}
    \begin{tabular}{lc|ccccc}
        \toprule
         \textbf{Setup} & \textbf{Expansion} & \textbf{mIoU $\uparrow$} & \textbf{FID $\downarrow$} & \textbf{SSIM $\uparrow$} & \textbf{PSNR $\uparrow$}\\ 
        \midrule
        \multirow{2}*{Indoor} &  1.17x & \textbf{0.86} & \textbf{43.48} & \textbf{0.94} & \textbf{23.06}\\
        {} & 1.40x & 0.55 & 75.42 & 0.84 & 17.33  \\
        % \addlinespace[0.1cm]
        {} & 1.75x & 0.34 & 110.01 & 0.78 & 14.90 \\
        \midrule
         % {} & \textbf{Outdoor Images Masking} & \textbf{Jaccard Index (mIoU) $\uparrow$} & \textbf{FID $\downarrow$} & \textbf{SSIM $\uparrow$} & \textbf{PSNR $\uparrow$} & \textbf{MSE $\downarrow$}  \\
        % \midrule
        \multirow{2}*{Outdoor}  & 1.17x & \textbf{0.90} & \textbf{42.63} & \textbf{0.94} & \textbf{25.96} \\
        {} & 1.40x & 0.73 & 73.03 & 0.86 & 21.39  \\
        % \addlinespace[0.1cm]
        {} & 1.75x &  0.57 & 118.56 & 0.79 & 18.80 \\
        \bottomrule
    \end{tabular}
  \label{tab:semantic}
  \vspace{-3mm}
\end{table}

\begin{table}[h]
  \vspace{3mm}
  \centering
  \caption{Results for increasing the FoV in  Binary images from Outdoor environment}
    \begin{tabular}{c|ccccc}
        \toprule
         \textbf{Expansion} & \textbf{mIoU $\uparrow$} & \textbf{FID $\downarrow$} & \textbf{SSIM $\uparrow$} & \textbf{PSNR $\uparrow$}\\ 
        \midrule
        1.17x & \textbf{0.90} &\textbf{ 51.87} & \textbf{0.95} & \textbf{30.36} \\
        1.40x & 0.78 & 88.44 & 0.76 & 22.05 \\
        % \addlinespace[0.1cm]
        1.75x & 0.64 & 120.94 & 0.56 & 17.81 \\
        \bottomrule
    \end{tabular}
  \label{tab:binary}
  \vspace{-3mm}
\end{table}

\begin{figure}[t]
\vspace{3mm}
  \centering
  \includegraphics[width=0.99\linewidth]{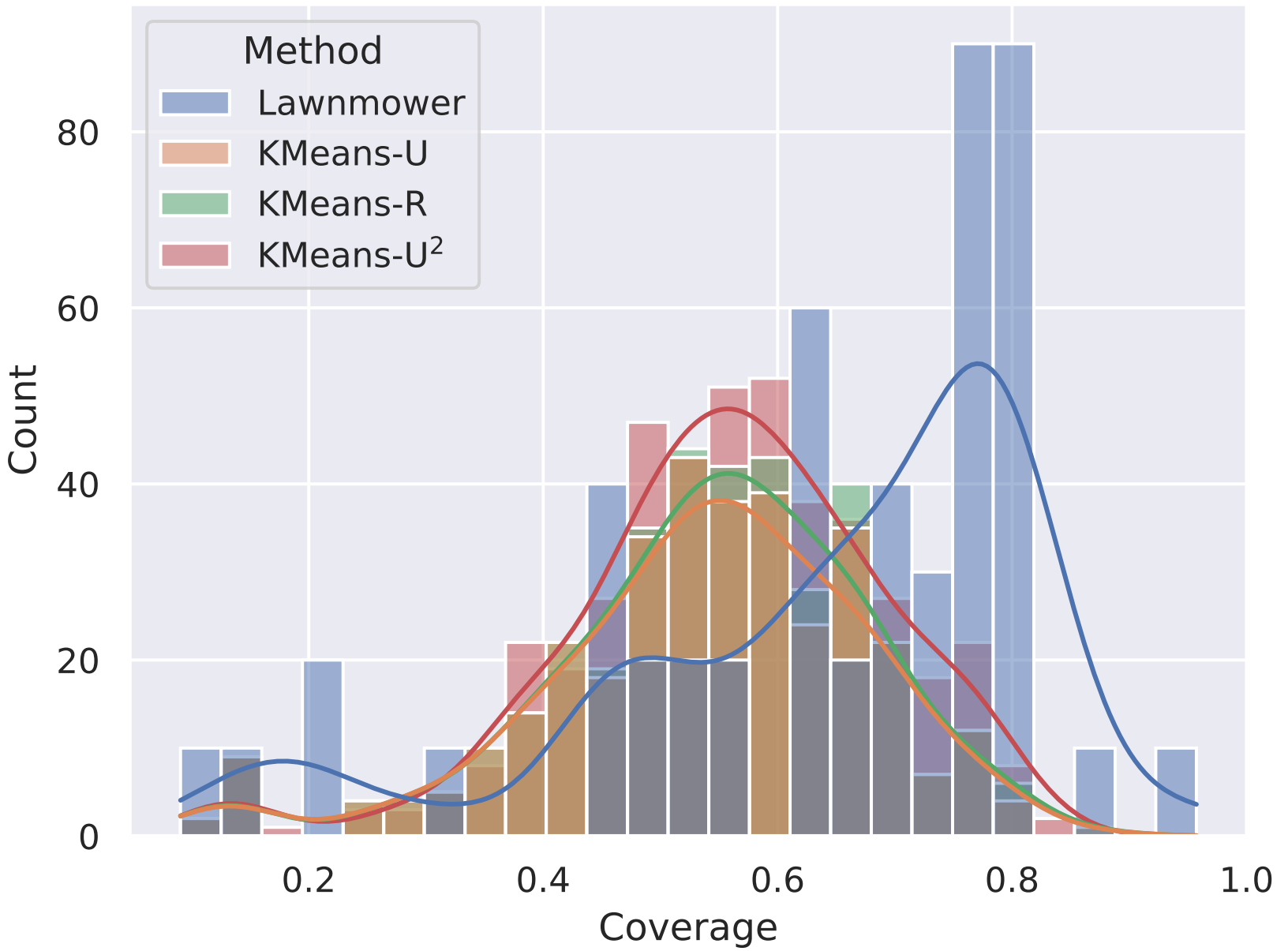}
  \caption{Comparison between the multi-agent exploration algorithms to reach at least $95\%$ accuracy in prediction of the unexplored map.}
  \label{fig:multiagent_explore_histplot}
  \vspace{-5mm}
\end{figure}

\subsection{Multi-Agent Uncertainty Guided Exploration}
\label{subsec:eval_multi_agent_exploration}
\vspace{-1mm}
For this task, we use 3-channel semantic map representations, consisting of free, occupied, and out-of-boundary regions, using color-to-label matching on the MAE prediction for labeling, as described in Section~\ref{subsec:method_fov_expansion}. In our experiments, we utilize 50 living room scenes from the ProcTHOR~\cite{deitke2022️} framework. We convert their ground truth semantic segmentation maps for the 3-class labeling. These labeled images are transformed into 3-channel RGB images, with free, occupied, and out-of-boundary regions represented by green, red, and blue colors, respectively.

We use $n=3$ aerial robots and conduct 10 experiments in each room, resulting in 500 runs total for each algorithm. We select the initial positions of the robots randomly. We assume that the area of the room to be explored is known beforehand and that the robots fly at a height taller than the obstacles and do not collide with each other. Each map is represented as an image with dimensions $224 \times 224$ pixels, and each robot can observe an area of size $48\times48$ pixels around it. We treat this as a centralized task, and the observations from all the robots are combined for decision-making.

\textbf{Results:} To compare the methods, we look at the distribution of coverage to reach at least 95\% accuracy in predicting the whole map given the partial observation. We visualize our findings in Fig.~\ref{fig:multiagent_explore_histplot}. As shown, most runs with Lawnmower need to cover around 75-85\% of the area. This happens due to the naive movement pattern of the robots with Lawnmower and thus the robots do not benefit from the inpainting capability of MAE. All KMeans algorithms, on the other hand, are able to take advantage of it and therefore most runs with them need only 50-60\% coverage to reach the same accuracy. KMeans-U$^2$ is especially denser here as it guides the robots to areas with uncertainty, reducing the chances of incorrect predictions. We note that some heavy-tailed behavior is observed in these plots as some rooms are very simple, and a few predictions may be enough to make good predictions in them. Additionally, spawning robots at the start of the scanlines with Lawnmower places them far apart initially, an advantage other algorithms do not enjoy. This results in Lawnmower sometimes getting better accuracy with less coverage in a simple environment.

These findings highlight an intriguing observation about regions with regularly shaped objects: most shapes can be reasonably inferred by looking only at a part of them. As a result, areas with such objects may not be as beneficial for exploration after partial observation, as the large unexplored regions are. KMeans-U$^2$ performs better as it prioritizes exploring unexplored regions only when it can make a confident (low variance) estimate about objects based on the partial view. The effectiveness of this approach hinges on a prediction model's accuracy in making precise predictions, a task which our experiments have shown MAE excels at. 
\vspace{-1mm}

\subsection{Navigation with prediction}
\label{subsec:eva_navigation}
For this application, we use a setup similar to the previous task. Specifically, we represent the occupancy map as a 3-channel semantic map and use color-to-label matching on the MAE predictions. We select 5 large living room scenes from ProcTHOR~\cite{deitke2022️} and choose 20 start-goal pairs on them, located far apart. The map size, robot's field of view, and prediction input are similar to Section~\ref{subsec:eval_multi_agent_exploration}. The unseen area is predicted by MAE. Using the predicted segmented map, we generate a costmap and use A$^*$ path planning algorithm for navigation. This process is repeated till the robot reaches the goal.

\textbf{Results:} 
Across twenty generated paths, our method takes on an average $10.5$ steps with a standard deviation of $2.9$ steps, whereas the traditional method takes $21.6$ steps with a standard deviation of $7.3$ steps. It is worth noting that with predictions, the larger frontier helps the robot estimate the shape of an obstacle beforehand, based on partial views, which leads to a reduction of $48\%$ in the total number resulting in efficient navigation.

\section{Limitations and Future Work}
In this work we show how MAE, a self-supervised network, pre-trained on first-person-view images can be applied to prediction-augmented robotic tasks reliant on top-down maps, without any fine-tuning. Our experiments show its applicability across various robotic tasks, involving different types of input modalities. A key takeaway from our work is that such models are capable of reasoning about regular geometric shapes and directly benefit robots in an environment filled with such patterns. We hope our analysis paves the way for further studies and the development of applications based on such powerful models. 

Our work focuses on the efficacy of a pre-trained model and is especially suitable for applications suffering from a lack of training data. We expect improvement in results with task-specific fine-tuning in future works. A drawback of MAE is that it requires the mask to be composed of square patches. While this can be attained with some innovative engineering when the required masks are irregular, other models supporting unrestricted masking might be more suited for this job. Whether they are able to retain the benefits of MAE or not will be explored in our future work.

{\small
\bibliographystyle{ieee_fullname}
\bibliography{egbib}
}

\end{document}